\documentclass[10pt,letterpaper,logo,twocolumn,teaser]{planstyle}
\usepackage[T1]{fontenc}
\usepackage{microtype}           
\usepackage{nicefrac}            
\usepackage{xspace}              
\usepackage{amsmath, amssymb, amsfonts, amsthm, mathtools, mathrsfs}
\usepackage{dsfont}              
\usepackage{fdsymbol}            

\usepackage{booktabs}            
\usepackage{nicematrix}         
\usepackage{multirow}
\usepackage{makecell}
\usepackage{bigstrut}
\usepackage{enumitem}            
\setitemize{label=\textbullet, leftmargin=*, nolistsep}
\usepackage{graphicx}
\usepackage{adjustbox}           
\usepackage{float}               
\usepackage{wrapfig}             
\usepackage{subcaption}          
\expandafter\def\csname ver@subfig.sty\endcsname{}  
\usepackage[dvipsnames]{xcolor}
\usepackage{color}
\usepackage{fontawesome5}        
\usepackage{pifont}              
\usepackage{hyperref}            
\hypersetup{colorlinks=true, citecolor=LightBlue}
\usepackage[all]{hypcap}         
\usepackage{cleveref}            
\usepackage{url}                 
\usepackage{algorithm}
\usepackage{algorithmicx}
\usepackage{algpseudocode}
\usepackage[normalem]{ulem}      
\usepackage{lipsum}              
\usepackage{rotating}            
\usepackage{stackengine}         
\usepackage{caption}             
\usepackage{listings}            
\usepackage{soul}                
\usepackage{gradient-text}       
\usepackage{pgfplots}            
\pgfplotsset{compat=newest}
\usepackage{pgfplotstable}       
\usepackage{tikz}                
\usepackage{svg}                 

\usepackage[comma,numbers,sort,compress]{natbib}

\hypersetup{
    colorlinks = true,
    citecolor = {YaleBlue},
}
\definecolor{demphcolor}{RGB}{125,125,125}             
\tcbuselibrary{breakable, skins}
\newtcolorbox{planbox}[1]{
  enhanced,
  breakable,
  colback=white,
  colframe=SYSURed!80,
  coltitle=SYSUGreen,
  fonttitle=\bfseries\sffamily,
  title=#1,
  titlerule=0.8pt,
  boxrule=1pt,
  left=3mm, right=3mm, top=2mm, bottom=2mm,
  boxsep=1mm,
  before upper=\smallskip,
}

\newcommand{\ie}{\textit{i.e.},\xspace}      
\newcommand{\eg}{\textit{e.g.},\xspace}      
\newcommand{\etc}{\textit{etc}.\xspace}      

\crefname{equation}{Eq.}{Eqs.}
\crefformat{section}{\S#2#1#3}
\crefformat{subsection}{\S#2#1#3}
\crefformat{subsubsection}{\S#2#1#3}
\crefrangeformat{section}{\S\S#3#1#4 to~#5#2#6}
\crefmultiformat{section}{\S\S#2#1#3}{ and~#2#1#3}{, #2#1#3}{ and~#2#1#3}

\PassOptionsToPackage{pagebackref,breaklinks,colorlinks,allcolors=cvprblue}{hyperref}

\usepackage[T1]{fontenc}
\usepackage{microtype}           
\usepackage{nicefrac}            
\usepackage{xspace}              
\usepackage{amsmath, amssymb, amsfonts, amsthm, mathtools, mathrsfs}
\usepackage{dsfont}              
\usepackage{fdsymbol}            

\usepackage{booktabs}            
\usepackage{nicematrix}         
\usepackage{multirow}
\usepackage{makecell}
\usepackage{bigstrut}
\usepackage{pifont}
\usepackage{enumitem}            
\setitemize{label=\textbullet, leftmargin=*, nolistsep}
\usepackage{graphicx}
\usepackage{adjustbox}           
\usepackage{float}               
\usepackage{wrapfig}             
\usepackage{subcaption}          
\expandafter\def\csname ver@subfig.sty\endcsname{}  
\usepackage{color}
\usepackage{fontawesome5}        
\usepackage{pifont}              
\usepackage{hyperref}            
\hypersetup{colorlinks=true, citecolor=LightBlue}
\usepackage[all]{hypcap}         
\usepackage{url}                 
\usepackage{algorithm}
\usepackage{algorithmicx}
\usepackage{algpseudocode}
\usepackage[normalem]{ulem}      
\usepackage{lipsum}              
\usepackage{rotating}            
\usepackage{stackengine}         
\usepackage{caption}             
\usepackage{listings}            
\usepackage{soul}                
\usepackage{gradient-text}       
\usepackage{pgfplots}            
\pgfplotsset{compat=newest}
\usepackage{pgfplotstable}       
\usepackage{tikz}                
\usepackage{svg}                 
\usepackage{amsmath,amssymb}
\usepackage{tcolorbox}
\tcbuselibrary{breakable}

\definecolor{LightBlue}{rgb}{0.68, 0.85, 0.9}
\definecolor{amberyellow}{rgb}{1.0, 0.75, 0.0}
\definecolor{DarkGreen}{rgb}{0.0, 0.5, 0.0}
\definecolor{DarkBlue}{rgb}{0,0,205}

\title{PhyDetEx: Detecting and Explaining the Physical Plausibility of T2V Models}

\author{
\begin{tabular}{c}
\textbf{Zeqing Wang\(^{1,3}\), Keze Wang\(^{1\dag}\), Lei Zhang\(^{2,3\dag}\)}
\end{tabular}
}

\affil{\textcolor{SYSUGreen}{$^1$Sun Yat-sen University, $^2$Hong Kong Polytechnic University
, $^3$OPPO Research Institute}}

\correspondingauthor{
\(^\dag\) Corresponding Author
}

\begin{document}

\begin{abstract}
Driven by the growing capacity and training scale, Text-to-Video (T2V) generation models have recently achieved substantial progress in video quality, length, and instruction-following capability. However, whether these models can understand physics and generate physically plausible videos remains a question. While Vision-Language Models (VLMs) have been widely used as general-purpose evaluators in various applications, they struggle to identify the physically impossible content from generated videos. 
To investigate this issue, we construct a \textbf{PID} (\textbf{P}hysical \textbf{I}mplausibility \textbf{D}etection) dataset, which consists of a \textit{test split} of 500 manually annotated videos and a \textit{train split} of 2,588 paired videos, where each implausible video is generated by carefully rewriting the caption of its corresponding real-world video to induce T2V models producing physically implausible content.
With the constructed dataset, we introduce a lightweight fine-tuning approach, enabling VLMs to not only detect physically implausible events but also generate textual explanations on the violated physical principles. Taking the fine-tuned VLM as a physical plausibility detector and explainer, namely \textbf{PhyDetEx}, we benchmark a series of state-of-the-art T2V models to assess their adherence to physical laws. Our findings show that although recent T2V models have made notable progress toward generating physically plausible content, understanding and adhering to physical laws remains a challenging issue, especially for open-source models.
\vspace{1mm}

\textcolor{DarkBlue}{\faGlobe} \url{https://github.com/Zeqing-Wang/PhyDetEx}
\vspace{-0.3cm}
\end{abstract}

\maketitle

\section{Introduction}
\label{sec:intro}
Recent advances in Text-to-Video (T2V) generation models have significantly improved the visual quality, temporal consistency, \etc, of the generated videos~\cite{Veo3Google,wan2025,kong2024hunyuanvideo,keling,ma2025stepvideot2vtechnicalreportpractice,yang2024cogvideox}. With the rapid scaling of model capacity and multimodal training data, these models are increasingly viewed as promising candidates toward realising a world model~\cite{phyworld,huang2025vid2worldcraftingvideodiffusion,wang2025wisa}, which can describe the dynamics and causality of the physical world. However, despite impressive progress, existing T2V models often produce physically implausible content that contradicts basic real-world principles~\cite{wang2025generated, bansal2024videophy, bansal2025videophy}. Detecting and explaining these violations is crucial to measure how well current T2V models can understand the physical world. 

\begin{figure}[t]
  \centering
  \captionsetup{skip=4pt}
   \includegraphics[width=1\linewidth]{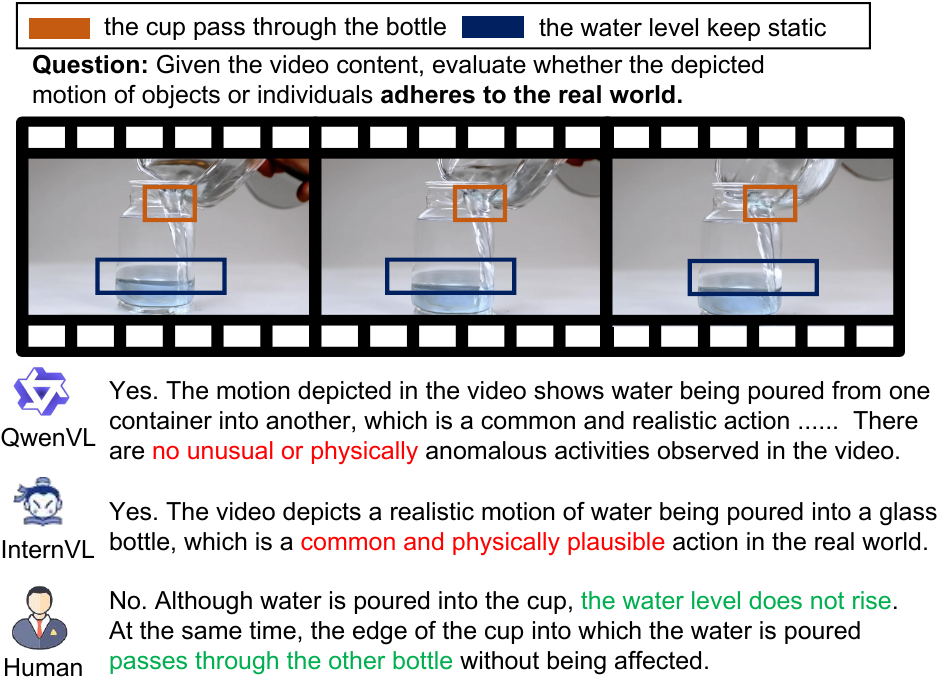}
   \caption{Illustration of the physical implausibility detection task.
Given a video where water is poured into a bottle, but the water level remains static and the cup passes through the bottle, humans can easily identify the violation of physical laws. However, current powerful VLMs (e.g., QwenVL and InternVL) incorrectly judge the motion as physically plausible, highlighting their difficulty in recognising implausible dynamics.}
   \label{fig:strange}
\end{figure}

Several studies attempt to assess the physical plausibility of generated videos using either pre-defined prompts~\cite{bansal2024videophy,bansal2025videophy,xue2025phyt2v,meng2024towards} or detectors trained with extensive human annotations~\cite{LiFT,bansal2024videophy,bansal2025videophy}. However, the first kind of methods are inherently restricted by their reliance on carefully designed prompts, while the second suffer from poor transferability and limited interpretability, which often produce only coarse plausibility scores without revealing why physical laws are violated. Consequently, these approaches remain narrow in scope and generality, underscoring the need for a more generalizable and interpretable evaluation framework that can handle a wide range of generated videos.

A natural approach to evaluate the physical plausibility of the generated videos with an explanation is to leverage Vision-Language Models (VLMs), which have achieved remarkable success across diverse perception and reasoning tasks~\cite{comanici2025gemini25pushingfrontier,Qwen2.5-VL,chen2024internvl,chen2024far,liu2023llava}. VLMs are trained on massive visual–text corpora~\cite{cui2025comprehensive} and possess strong generalization and interpretability through textual reasoning~\cite{jiang2025vlm,zhang2025mavis}. It is intuitive to use them as general evaluators to identify physically implausible content in videos. However, recent evidence shows that even the most capable VLMs struggle to recognise physically implausible content generated by T2V models~\cite{bai2025impossible, wang2025generated}. As shown in Fig.~\ref{fig:strange}, VLMs often fail to detect implausible content that is easy for humans to perceive. This motivates us to investigate whether we can endow VLMs, which have shown strong capability in many downstream visual tasks, with the ability to detect and explain the physical implausibility in generated videos. 

To answer the above question, we first conduct a preliminary experiment by prompting several state-of-the-art VLMs to detect the physical pausibility in videos with varying degrees of prior information, without any additional training. We observe that their detection accuracy changes substantially depending on the given priors, suggesting that VLMs may possess certain physical knowledge, yet they struggle to apply this knowledge to unseen implausible cases without proper training. 
Therefore, we first construct a \textbf{P}hysical \textbf{I}mplausibility \textbf{D}etection (\textbf{PID}) dataset, which includes a \textit{train split} and a \textit{test split}. 
The \textbf{train split} consists of 2,588 paired videos, where each pair consists of a physically plausible (positive) video and an implausible (negative) counterpart. The implausible videos are generated by carefully rewriting captions to induce T2V models to produce physically impossible videos. 
The \textbf{test split} consists of 500 videos, including 250 physically implausible videos with manually annotated content of physical violations and 250 physically plausible counterparts. The implausible videos are collected from various T2V models that unintentionally generate abnormal content under normal prompts, while the plausible ones are drawn from both real-world videos and high-quality T2V generations.

Using the PID train split, we propose a lightweight fine-tuning approach that requires only minimal computational cost (via LoRA-based parameter-efficient tuning). The resulting model, namely \textbf{PhyDetEx} (\textbf{Phy}sical Plausibility \textbf{Det}ector and \textbf{Ex}plainer), empowers VLMs to (i) accurately identify physically implausible events and (ii) produce textual explanations of the violated physical principles. Our experiments demonstrate that PhyDetEx achieves significant improvements in detecting physical implausibility, validating that VLMs indeed possess an implicit understanding of physics that can be unlocked with a small number of contrastive examples.
Finally, we apply PhyDetEx as an evaluation tool to benchmark the physical plausibility of current T2V models. Our results reveal that while recent closed-source models have made remarkable strides toward generating physically consistent videos, understanding and adhering to physical laws remain a substantial challenge, especially for open-source models. These findings highlight both the progress and the limitations of current T2V systems in genuine world modeling.

In summary, our contributions are as follows. First, we introduce \textbf{PID}, a comprehensive benchmark for physical implausibility detection of T2V models, which includes a \textit{test split} of 500 annotated videos (250 physically plausible and 250 implausible) and a \textit{train split} of 2,588 plausible/implausible paired videos. Second,  we reveal the reasons behind VLMs' poor performance on physical implausibility based on comprehensive experiments. Third, we propose a lightweight contrastive fine-tuning strategy, which enables VLMs to detect and explain physical implausibility in generated videos. Finally, we employ the fine-tuned PhyDetEx to evaluate the physical plausibility of mainstream T2V models, providing new insights on their progress and challenges toward world modeling.

\section{Related Work}
\label{sec:related_work}

\noindent\textbf{Text-to-Video (T2V) Models.} Recent large-scale T2V models have demonstrated significantly improved video fidelity, motion coherence, and instruction following capability~\cite{kong2024hunyuanvideo,keling,li2024sora,minmaxhailuo,PikaLab,wan2025,yang2024cogvideox,xiao2025captain}, making them a promising path toward building world models~\cite{cho2024sora,qin2025worldsimbench,xiang2024pandora} that can capture the causal dynamics of the real world. However, despite the impressive improvement in video quality, these models often violate basic physical laws~\cite{zhu2024sora,phyworld,zhang2025videorepa,wang2025videoverse}, such as object penetration, momentum inconsistency, or energy non-conservation. Such implausible generations even under semantically normal prompts are typically unacceptable for practical applications, which necessites the need to evaluate the physical plausibility beyond visual quality.

\noindent\textbf{Physical Implausibility Detection.} Given the prevalence of physically implausible content in T2V outputs, several approaches have been proposed for detecting it. One line of work trains end-to-end detectors with annotated datasets~\cite{bansal2024videophy,bansal2025videophy,duan2025worldscore,LiFT,he-etal-2024-videoscore}, assigning a physical plausibility score to generated videos. Another direction defines a set of rule-based prompts with known physical attributes and checks whether the generated content satisfies these predefined conditions~\cite{zheng2025vbench,xue2025phyt2v,chen2025t2vworldbenchbenchmarkevaluatingworld,meng2025towards,wang2025videoverse}. However, both paradigms face key limitations: end-to-end detectors often lack interpretability, while rule-based methods are restricted to predefined prompt scenarios. Hence, a more generalizable and interpretable method is required to measure models' adherence to physical laws. In contrast, Vision-Language Models~\cite{Qwen2.5-VL,chen2024internvl,chen2024far,liu2023llava}, which are trained on massive multimodal data~\cite{cui2025comprehensive,lin2014microsoft} and are capable of producing textual explanations~\cite{scienceQA,AOKVQA,vipergpt,visualprogramming}, offer a promising alternative for interpretable and generalizable detection.

\noindent\textbf{Vision-Language Models (VLMs).} VLMs have achieved remarkable progress in recent years~\cite{Qwen2.5-VL,chen2024far,liu2023llava,li2024llava}, excelling not only in traditional vision tasks such as visual question answering~\cite{AOKVQA,gqa,vqav2} and captioning~\cite{vinyals2015show} but also in complex reasoning tasks like visual grounding and generation~\cite{OMGLLaVA,zhan-etal-2024-anygpt,deng2025bagel}. Nonetheless, prior studies reveal that even strong VLMs struggle to detect physically implausible events in generated content~\cite{bai2025impossible, wang2025generated}. Impossible Videos~\cite{bai2025impossible} is the first work to evaluate VLMs' understanding of such implausibility, establishing a benchmark to assess their detection ability. However, it focuses solely on videos generated by implausible prompts and cannot reflect real-world T2V failures caused by normal prompts. These limitations motivate us to explore how to equip VLMs with the capability to not only detect physically implausible events and explain the violated physical principles, but also identify real-world T2V failures, thereby benchmarking the physical plausibility of current T2V models.

\section{Preliminary Investigation}
\label{sec:pre}
As introduced in Sec.~\ref{sec:intro}, prior studies~\cite{bai2025impossible,wang2025generated} have shown that even powerful VLMs struggle to detect physically impossible or implausible content generated by T2V models. However, it is unclear whether these failures stem from a lack of understanding of physical plausibility, or are caused by the biases caused by training predominantly on physically plausible real-world data.
To find out the reason and provide guidance for future improvements, we conduct a series of preliminary analyses. Specifically, we evaluate VLMs under three different prompt configurations with progressively increasing prior information based on the Impossible Videos benchmark~\cite{bai2025impossible}. Since all implausible videos in Impossible Videos are generated, we define three conditions:
\textbf{(C1)} the model is not informed that the video is generated (our primary evaluation setting);
\textbf{(C2)} the VLMs are told that the video may be generated; and
\textbf{(C3)} the VLMs are explicitly told that the video is generated by a T2V model.

\begin{figure*}[t]
  \centering
  \captionsetup{skip=4pt}
   \includegraphics[width=1\linewidth]{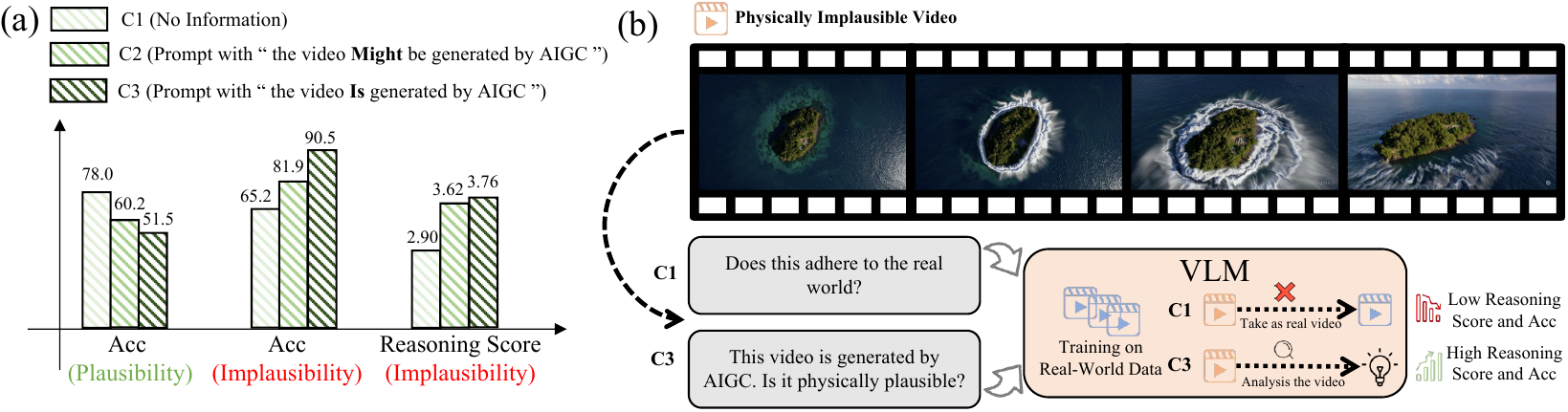}
   \caption{(a) Results of preliminary experiments in the ImpossibleVideos under three prompting conditions (\textbf{C1–C3}) based on InternVL2.5 26B. As the prompt provides progressively stronger hints that the video may be generated by an AIGC model, the VLM achieve notably higher accuracy in detecting \textit{physically implausible} videos and generate more accurate reasoning (higher reasoning scores). However, their accuracy on \textit{physically plausible} videos decreases accordingly. These trends indicate that VLMs possess an implicit understanding of physical plausibility, yet their judgments are strongly biased by the type of input video they consider (real or generated). (b) Since most of its training data comes from the real world, VLM tends to assume that the input is real-world videos without providing any information.}
   \label{fig:pre_exp_fig}
\end{figure*}

As shown in Fig.~\ref{fig:pre_exp_fig} (a), the detection performance varies substantially across these different conditions. The accuracy of identifying physically implausible content consistently increases from \textbf{C1} to \textbf{C3}, whereas the accuracy for physically plausible videos decreases accordingly. Furthermore, VLMs not only improve in detection accuracy but also produce \textbf{more precise explanations} for the implausible events. These findings reveal that VLMs inherently possess a latent understanding of physical laws, but their reasoning is heavily influenced by the assumed authenticity of the type of input video. As shown in Fig.~\ref{fig:pre_exp_fig} (b), when VLMs presume the video is real (as most of their training data are), they tend to overlook violations of physical plausibility. Conversely, when explicitly informed that the video is generated, they adopt a more critical stance and successfully identify physical implausibility. We provide more preliminary experiments with different VLMs and details in Sec~\ref{sec:exp} and \textbf{Supplemental Material A}.

\section{Methodology}
\label{sec:method}

\begin{figure*}[t]
  \centering
  \captionsetup{skip=4pt}
   \includegraphics[width=1\linewidth]{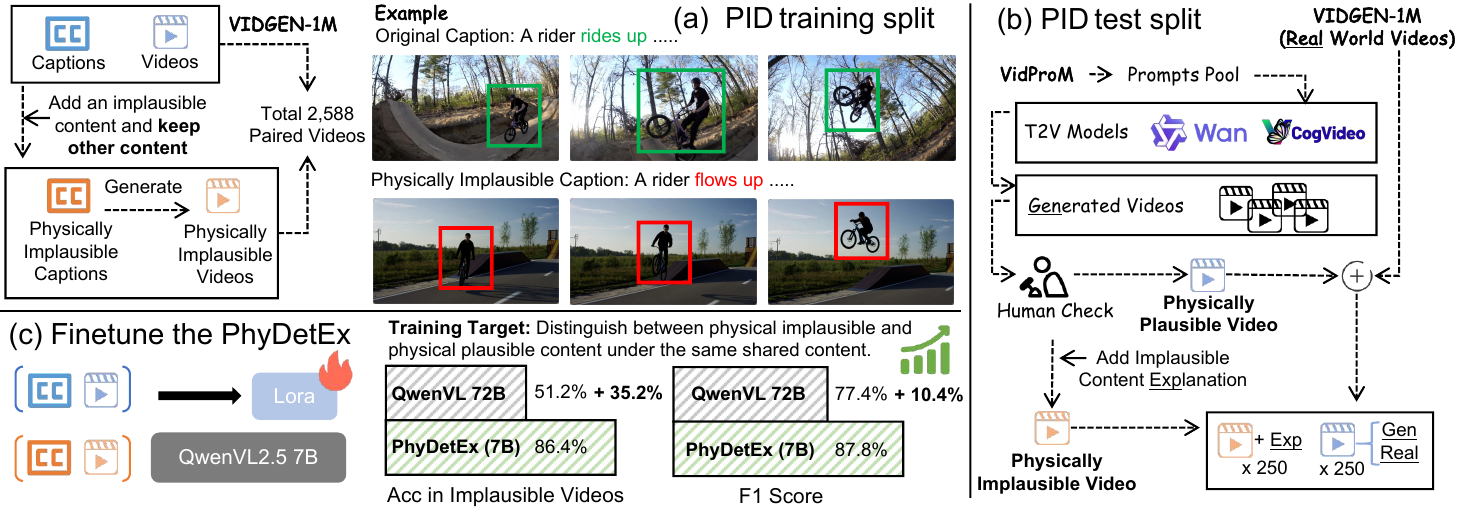}
   \caption{Overview of the construction pipeline of the PID dataset and the training process of PhyDetEx. (a) The PID training split. The training split includes 2,588 paired videos, where each implausible video is generated by rewriting the caption of a real-world video to describe an implausible event while keeping other content unchanged. (b) The PID test split. The test set consists of physically implausible and physically plausible videos. The implausible subset is collected from videos generated by multiple T2V models based on physically plausible prompts, where human annotators identify implausible events and provide textual explanations. The plausible subset combines generated and real-world videos verified to contain no physical violations, thereby eliminating the shortcut of distinguishing between generated and real videos. (c) Training the PhyDetEx. Using the PID training split, we finetune the base VLM via LoRA adaptation to distinguish between physically plausible and implausible events within the same video contexts. The resulting model, PhyDetEx, achieves substantial improvements in detecting physically implausible content.}
   \label{fig:pid_and_training}
\end{figure*}

\subsection{Construction of PID}
Building upon our preliminary findings, we observe that VLMs indeed possess the capability to understand physically implausible content in generated videos. However, because their training data predominantly consist of physically plausible, real-world content, they tend to assume that any given video is real, and thus free of anomalies. Conversely, they often over-associate generated videos with implausibility. To mitigate this bias, we build the \textbf{P}hysical \textbf{I}mplausibility \textbf{D}etection (PID) dataset to help VLMs detect implausible content within generated data rather than merely distinguishing between real and generated videos.

\vspace{+1mm}
\noindent\textbf{Training Split}. As shown in Fig.~\ref{fig:pid_and_training}(a), the training split of PID is built from the real-world T2V dataset, VIDGEN-1M~\cite{tan2024vdgen-1m}. The goal of PID's training split is twofold:
(1) it should not require a massive scale, since VLMs already exhibit a latent understanding of physical plausibility; and
(2) it should explicitly expose the models to physical implausibility patterns that are absent from their pretraining data, but the overall content of the data should still be highly relevant to the real world, thereby breaking the shortcut that equates ``generated videos'' with ``physically implausibile videos''.

To construct the dataset, we first select videos from VIDGEN-1M that involve clear physical interactions with their corresponding captions. We then employ an LLM to rewrite each caption such that it contains a physically implausible description while preserving all other contextual elements. A T2V model is then used to generate videos based on these rewritten captions. This process produces video pairs where the physically plausible videos originate from VIDGEN-1M, and the corresponding physically implausible videos are generated from the modified caption.
Based on our preliminary investigation, VLMs can reliably identify implausible content when provided with prior information. We then instruct the VLM that the videos are generated by a T2V model and use it to filter out pairs where the physically implausible content is accurately reflected in the generated video. Through this process, we obtain 2,588 paired videos.

This construction pipeline offers two major advantages. First, by generating physically implausible videos from the real-world ones, the PID training split closely mirrors the true distribution of physical content and avoids biases introduced by hand-defined ``implausibility classes''. Second, as illustrated in Fig.~\ref{fig:pid_and_training}(a), during caption rewriting, we modify only the physical aspect of the scene while keeping all other elements unchanged. Consequently, each video pair differs only in physical plausibility, forming a contrastive structure that isolates the target concept. This ensures that the generated videos remain semantically similar to their real counterparts, reducing the domain gap from the VLMs' training data and eliminating the shortcut of detecting the physical implausibility. The advantages of such a design are further analysed in \textbf{Supplemental Material B}.

\vspace{+1mm}
\noindent\textbf{Test Split}. To verify whether the model truly acquires the ability to detect physical implausiblity, we introduce the test split of our PID.
While Impossible Videos~\cite{bai2025impossible} offers an initial benchmark to evaluate the physical implausibility detectiion ability of VLMs, it suffers from two key limitations.
First, most of its videos are derived from explicitly physically implausible prompts, which diverge from realistic scenarios, where users typically input plausible prompts, but T2V models still produce physically implausible videos.
Second, the physically plausible videos of Impossible Videos consist entirely of real-world videos, creating a shortcut: VLMs may simply distinguish between generated and real videos instead of genuinely reasoning about physical plausibility.

To address these issues, we construct the PID test split, specifically designed to remove such shortcuts and more faithfully assess VLMs' understanding of physical laws. As illustrated in Fig.~\ref{fig:pid_and_training}(b), the construction of test split involves several stages.
We first sample prompts from VidProM~\cite{wang2024vidprom}, a large-scale prompts dataset crawled from the real user community, filtering out those containing inherently unrealistic or science fictional elements (\eg, cartoon, futuristic, magical), as well as prompts lacking physical interactions (\eg, static scenes). 
Using these filtered prompts, we generate videos via multiple T2V models and manually identify those exhibiting physical implausibility, accompanied by textual explanations describing the physically implausible content.
For the physically plausible subset, to eliminate the ``generated vs. real'' shortcut, we deliberately avoid relying solely on real-world videos. Instead, we combine physically plausible generated videos with real-world videos from VIDGEN-1M~\cite{tan2024vdgen-1m}, filtered to be under 10 seconds and contain clear physical interactions. Following this pipeline, the PID test split comprises 250 physically plausible and 250 physically implausible videos, resulting in a total of 500 balanced test samples.

We provide detailed statistics and more details of the construction pipeline of our proposed PID train split and test split in \textbf{Supplemental Material C}.


\subsection{Tuning the VLM}
With the constructed PID dataset, our objective is to tune the VLM so that it can accurately detect \textit{physically implausible} content in the generated videos. As discussed in Sec.~\ref{sec:pre}, VLMs tend to treat ``real'' as ``physically plausible'' and ``generated'' as ``physically implausible''. Our tuning should remove this shortcut and enable the VLMs to reason about whether a video contains physically implausible content.

\vspace{+1mm}
\noindent\textbf{Notation.} 
Denote by $v_r$ a real-world video and $t_r$ its corresponding caption with physically plausible events.  
We first rewrite $t_r$ using an LLM, producing a modified caption $t_g$ where the physically plausible event is changed to a physically implausible event, while keeping all other contextual elements the same:
\begin{equation}
t_g = \mathcal{R}_{\text{phys}}(t_r),
\end{equation}
where $\mathcal{R}_{\text{phys}}(\cdot)$ denotes the rewriting operation. Then, a physically implausible video $v_g$ is generated from the rewritten caption $t_g$ using a T2V model $\mathcal{G}$:
\begin{equation}
v_g = \mathcal{G}(t_g).
\end{equation}

The content in videos $v_r$ and $v_g$ can be partitioned into two parts, one part about the physically plausible/implausible event, and another shared part with the same content (\eg, environment, background objects, scene layout), denoted by $c_{\text{share}}$. Then, we can express the content of the two videos as follows:
\begin{equation}
v_r = (c_{\text{phys}}^r, c_{\text{share}}), \quad
v_g = (c_{\text{phys}}^g, c_{\text{share}}),
\end{equation}
where $c_{\text{phys}}^r$ and $c_{\text{phys}}^g$ represent the content of the physically plausible and implausible events in the real and generated videos, respectively. 
The resulting paired samples can then be represented as:
\begin{equation}
(v_r, t_r), \quad (v_g, t_g),
\end{equation}
where $(v_r, t_r)$ represents a physically plausible video from the real world, and $(v_g, t_g)$ represents the corresponding physically implausible video generated from the modified caption while preserving the other content.

\vspace{+1mm}
\noindent\textbf{Training Objective.}  
Given paired videos $(v_r, v_g)$, the goal of our training is to teach the VLM to correctly distinguish the content of each video as physically plausible or implausible. Let $f_\theta(\cdot)$ denote the prediction of a VLM model on the physical plausibility of a video:
\begin{equation}
y_r = f_\theta(v_r), \quad y_g = f_\theta(v_g),
\end{equation}
where $y_r, y_g \in \{ \text{physically plausible}, \text{physically implausible} \}$.  
The model is trained with the target such that, for each paired sample videos, the predicted labels should correctly reflect the plausibility of the physical content:
\begin{equation}
\begin{cases}
y_r = \text{physically plausible}, \\
y_g = \text{physically implausible}.
\end{cases}
\end{equation}

In other words, the VLM is encouraged to rely solely on the content $c_{\text{phys}}$ to make its decision, while ignoring the shared content $c_{\text{share}}$. This ensures that the model does not exploit shortcuts based on background, environment, or other unrelated content of the underlying physical event.  
There are two key motivations for this design. First, physical event is a highly diverse and context-dependent phenomenon that is difficult to categorize explicitly. By generating physically implausible samples through caption modifications of real-world videos, the resulting content remains within the distribution of real-world motions, preserving meaningful context dependencies; for instance, surfing motions only make sense in scenes containing water. Second, as discussed in Sec.~\ref{sec:pre}, our preliminary investigation indicates that VLMs may develop a shortcut correlating generated videos with physical implausibility and real-world videos with physical plausibility, rather than reasoning about the physics. 
Our design minimizes cues from generation artifacts and encourages the model to focus on the physical event itself, thereby mitigating such shortcuts.

\vspace{+1mm}
\noindent\textbf{Language Modeling Supervision.}  
Because the physically implausible videos $v_g$ are generated from rewritten captions $t_g$ with known physical implausibility, the model can be directly supervised to predict both the correct plausibility label and its corresponding reasoning. Similar to the current training objective of VLMs and LLMs, the entire training can be unified under a single language modelling (LM) objective.  
For each video–caption pair $(v, t)$, the VLM is required to generate a textual sequence $\mathbf{s} = [s_1, s_2, \dots, s_T]$, where the first token $s_1$ represents the physical plausibility judgment (\texttt{[YES]} or \texttt{[NO]}), and the remaining tokens $\{s_2, \dots, s_T\}$ form a natural-language explanation supporting this judgment.  
Given the visual and textual inputs, the model is trained autoregressively using the standard negative log-likelihood loss:
\begin{equation}
\mathcal{L}_{\text{LM}}(\mathbf{s} \mid v, t)
= - {\textstyle\sum_{u=1}^{T}} \log p_\theta(s_u \mid s_{<u}, v, t),
\label{eq:lm_loss}
\end{equation}
where $\theta$ denotes the trainable parameters of the VLM.  

This unified LM-based supervision activates the VLM's latent capability to detect physical implausibility and leverages its inherent strength in generating textual explanations, enabling the model to serve as both a reliable detector and an interpretable reasoner for physical plausibility.

\section{Experiments}
\label{sec:exp}

\subsection{Experiments Setup}
\textbf{Training Details}.
We adopt QwenVL2.5-7B as the base model and fine-tune it using LoRA on our PID training split for three epochs. The learning rate is set to 1e-4 with a cosine learning rate scheduler. We set the LoRA rank to 8 and apply it to all target modules to ensure comprehensive adaptation. To further validate the effectiveness of our PID train split and training strategy in enhancing the model's understanding of physical plausibility, we conduct additional analyses and experiments in \textbf{Supplemental Material B}.


\begin{table*}[t]
\centering
\captionsetup{skip=4pt}
\resizebox{2\columnwidth}{!}{
    \begin{tabular}{l cccc c cccc}
\toprule
\multirow{3}{*}{\textbf{Model}} 
& \multicolumn{4}{c}{\textbf{Impossible Video}} 
& \multicolumn{4}{c}{\textbf{PID Test}} \\

\cmidrule(lr){2-5} \cmidrule(lr){6-9}
& \textbf{Acc.} & \textbf{Acc.} & \textbf{F1} & \textbf{Reasoning} 
& \textbf{Acc.} & \textbf{Acc.} & \textbf{F1} & \textbf{Reasoning} \\[-2pt]
& \textbf{Impl.} & \textbf{Plaus.} & \textbf{Score} & \textbf{Score}
& \textbf{Impl.} & \textbf{Plaus.} & \textbf{Score} & \textbf{Score} \\
\midrule

Random Guess & 50.0 & 50.0 & 47.4 & -- & 50.0 & 50.0 & 50.0 & -- \\
\midrule
    InternVL2.5 4B~\cite{chen2024expanding} & 61.3  & 69.3  & 69.0    & 2.72  & 61.6  & 84.4  & 75.8  & 2.26 \\
    InternVL2.5 8B~\cite{chen2024expanding} & 52.3  & 76.9  & 71.2  & 2.32  & 53.2  & 90.8  & 76.4  & 2.13 \\
    InternVL2.5 26B~\cite{chen2024expanding} & 65.2  & 78.0    & 73.2  & 2.89  & 58.4  & 92.8  & 79.2  & 2.16 \\
    LLaVA One-Vision~\cite{li2024llava} & 29.3  & 81.5  & 68.0    & 1.70   & 36.8  & 94.0    & 73.1  & 1.75 \\
    MiniCPM-V 4.5 8B~\cite{yao2024minicpm} & 21.3  & \textbf{98.9}  & 75.0    & 1.91  & 10.4  & \textbf{97.6}  & 68.0    & 1.52 \\
    Qwen2.5VL 3B~\cite{Qwen2.5-VL} & 38.7  & 84.3  & 71.8  & 2.36  & 45.2  & 88.0    & 72.5  & 2.30 \\
    Qwen2.5VL 7B~\cite{Qwen2.5-VL} & 64.9  & 69.8  & 70.3  & 1.73  & 57.2  & 86.4  & 75.4  & 1.66 \\
    Qwen2.5VL 32B~\cite{Qwen2.5-VL} & 63.2  & 84.5  & 78.7  & 1.84  & 35.6  & 96.4  & 73.9  & 1.70 \\
    Qwen2.5VL 72B~\cite{Qwen2.5-VL} & 51.2  & 88.5  & 77.4  & 3.22  & 54.0  & 91.2  & 76.9  & 3.13 \\
    Qwen3VL 8B~\cite{Qwen-VL} & 47.3  & 97.8  & 81.1  & 3.20   & 51.6  & 93.6  & 77.4  & 3.13 \\
    Qwen3VL 30B A3~\cite{Qwen-VL} & 63.2  & 84.4  & 78.7  & 3.71  & 58.4  & 90.0    & 77.7  & 3.28 \\
    \midrule
    PhyDetEx (ours, 7B) & \textbf{86.4}  & 87.1  & \textbf{87.8}  & \textbf{4.20}   & \textbf{75.6}  & 89.2  & \textbf{83.5}  & \textbf{4.24}\\
    \bottomrule
\end{tabular}

    

}
\caption{Quantitative comparison of physical implausibility detection results across different VLMs on the Impossible Videos and our PID test split. The \textbf{bold} indicates the best performance.
We report four metrics for each model: accuracy on implausible videos (\textbf{Acc. Impl.}), accuracy on plausible videos (\textbf{Acc. Plaus.}), overall detection performance (\textbf{F1 Score}), and explanation quality (\textbf{Reasoning Score}). Our proposed PhyDetEx achieves the best performance on both datasets, significantly surpassing existing VLMs in F1 Score and Reasoning Score, demonstrating its superior capability in detecting and reasoning about the physical implausibility in generated videos.}
\label{tab:bench_res_pid_impossible}
\end{table*}

\vspace{+1mm}
\noindent\textbf{Evaluation Datasets}.
We evaluate our proposed PhyDetEx on two datasets: Impossible Videos~\cite{bai2025impossible} and our PID test split.
For Impossible Videos, to better align with our focus on physical plausibility detection, we select the videos under the \textbf{Physical Laws} category, comprising 535 physically implausible videos. For the physically plausible subset, we adopt all real-world videos, resulting in 650 physically plausible samples.
Our PID test split contains 250 physically implausible and 250 physically plausible videos. More details of our PID dataset are described in \textbf{Supplemental Material C}.

\vspace{+1mm}
\noindent\textbf{Evaluation Metrics}. We compare our PhyDetEx with several state-of-the-art VLMs using four metrics, specifically: (1) \textbf{Accuracy Implausible (Acc. Impl.)}: accuracy on physically implausible videos; (2) \textbf{Accuracy Plausible (Acc. Plaus.)}: accuracy on physically plausible videos; (3) the \textbf{F1 Score}: the overall detection performance balancing between the two accuracies; and (4) the \textbf{Reasoning Score}: the correctness and depth of the model's explanation for the detected physical implausibility.

The inclusion of the F1 Score is crucial, as many VLMs tend to exploit shortcuts, achieving high accuracy on the physically plausible class while failing on the other. For instance, MiniCPM-V 4.5 attains an exceptionally high Accuracy Plausible of 98.9\% on ImpossibleVideos, but this coincides with a very low Acc Implausible, indicating a strong bias toward assuming all videos are physically plausible. Thus, the F1 Score serves as a more reliable indicator of true detection capability. The Reasoning Score further evaluates whether the model genuinely understands the underlying physical inconsistency rather than relying on superficial correlations. Detailed metric definitions and evaluation protocols are provided in the \textbf{Supplemental Material D}.

\subsection{More Results of Preliminary Investigation}
\label{sub:detials_of_prelimeinary_exp}

In addition to InternVL2.5-26B (see  Fig.~\ref{fig:pre_exp_fig} (a)), we also evaluate several other VLMs, as shown in Tab.~\ref{tab:sub_pre_inves_res}. The results show that across nearly all VLMs, the output changes substantially when provided with different prompts under conditions C1–C3: \emph{Acc.~Impl} and \emph{Reasoning Score} increase significantly, while \emph{Acc.~Plaus} drops significantly. This suggests that VLMs tend to assume that a generated video is physically plausible, rather than leveraging their broad pretraining knowledge to assess the physical consistency of the video content. These observations further motivate the design and construction of our PID dataset.

\begin{table*}[t]
\centering
\captionsetup{skip=4pt}
\resizebox{2\columnwidth}{!}{
    \begin{tabular}{cccccccccc}
    \toprule
    \multirow{2}[4]{*}{Model} & \multicolumn{3}{c}{C1} & \multicolumn{3}{c}{C2} & \multicolumn{3}{c}{C3} \\
\cmidrule(lr){2-4} \cmidrule(lr){5-7} \cmidrule(lr){8-10}         & Acc. Impl & Acc. Plaus & Reasoning Score & Acc. Impl & Acc. Plaus & Reasoning Score & Acc. Impl & Acc. Plaus & Reasoning Score \\
    \midrule
    InternVL2.5 8B & \cellcolor[rgb]{ .851,  .882,  .949}52.3 & \cellcolor[rgb]{ .557,  .663,  .859}76.9 & \cellcolor[rgb]{ .851,  .882,  .949}2.32 & \cellcolor[rgb]{ .706,  .776,  .906}69.35 & \cellcolor[rgb]{ .706,  .776,  .906}74.77 & \cellcolor[rgb]{ .706,  .776,  .906}2.81 & \cellcolor[rgb]{ .557,  .663,  .859}87.29 & \cellcolor[rgb]{ .851,  .882,  .949}48.31 & \cellcolor[rgb]{ .557,  .663,  .859}3.11 \\
    InternVL2.5 26B & \cellcolor[rgb]{ .851,  .882,  .949}65.23 & \cellcolor[rgb]{ .557,  .663,  .859}78 & \cellcolor[rgb]{ .851,  .882,  .949}2.9 & \cellcolor[rgb]{ .706,  .776,  .906}81.87 & \cellcolor[rgb]{ .706,  .776,  .906}60.15 & \cellcolor[rgb]{ .706,  .776,  .906}3.62 & \cellcolor[rgb]{ .557,  .663,  .859}90.47 & \cellcolor[rgb]{ .851,  .882,  .949}51.54 & \cellcolor[rgb]{ .557,  .663,  .859}3.76 \\
    Qwen2.5VL 7B & \cellcolor[rgb]{ .851,  .882,  .949}53.08 & \cellcolor[rgb]{ .557,  .663,  .859}73.85 & \cellcolor[rgb]{ .706,  .776,  .906}1.73 & \cellcolor[rgb]{ .706,  .776,  .906}62.43 & \cellcolor[rgb]{ .706,  .776,  .906}71.85 & \cellcolor[rgb]{ .557,  .663,  .859}1.88 & \cellcolor[rgb]{ .557,  .663,  .859}64.86 & \cellcolor[rgb]{ .851,  .882,  .949}69.85 & \cellcolor[rgb]{ .851,  .882,  .949}1.69 \\
    Qwen2.5VL 32B & \cellcolor[rgb]{ .851,  .882,  .949}45.98 & \cellcolor[rgb]{ .557,  .663,  .859}94.62 & \cellcolor[rgb]{ .851,  .882,  .949}1.84 & \cellcolor[rgb]{ .706,  .776,  .906}52.52 & \cellcolor[rgb]{ .706,  .776,  .906}86.92 & \cellcolor[rgb]{ .706,  .776,  .906}2.04 & \cellcolor[rgb]{ .557,  .663,  .859}60.75 & \cellcolor[rgb]{ .851,  .882,  .949}82.46 & \cellcolor[rgb]{ .557,  .663,  .859}2.06 \\
    Qwen2.5VL 72B & \cellcolor[rgb]{ .706,  .776,  .906}51.21 & \cellcolor[rgb]{ .706,  .776,  .906}88.46 & \cellcolor[rgb]{ .851,  .882,  .949}3.22 & \cellcolor[rgb]{ .851,  .882,  .949}49.72 & \cellcolor[rgb]{ .557,  .663,  .859}89.08 & \cellcolor[rgb]{ .706,  .776,  .906}3.32 & \cellcolor[rgb]{ .557,  .663,  .859}56.07 & \cellcolor[rgb]{ .851,  .882,  .949}87.08 & \cellcolor[rgb]{ .557,  .663,  .859}3.38 \\
    \bottomrule
    \end{tabular}%

}
\caption{Preliminary investigation results across different VLMs under conditions C1–C3. For each model, we report \emph{Acc.~Impl}, \emph{Acc.~Plaus}, and \emph{Reasoning Score}. All models exhibit substantial performance shifts as the prompt condition changes.}
\label{tab:sub_pre_inves_res}
\end{table*}

\subsection{Experimental Results}
As shown in Tab.~\ref{tab:bench_res_pid_impossible}, on both Impossible Videos and our PID test split, most VLMs exhibit a notable bias when detecting physically implausible videos. For example, MiniCPM-V 4.5 achieves 97.6\% Acc. Plaus. but only 10.4\% Acc. Impl. on the PID test split. Similarly, LLaVA-One-Vision shows a comparable bias, leading to high Acc. Plaus. but low F1 Scores.
It is worth noting that, the recent Qwen3VL 8B and Qwen3VL 30B-A3 models, although having smaller model sizes than QwenVL2.5 32B and 72B, achieve the highest F1 Scores among all baselines on both our PID test and the Impossible Video benchmark. This observation suggests that current VLM models' capacity is sufficient for physical implausibility detection, but the lack of implausible content in pretraining data prevents them from fully unlocking their potential in detecting physical implausibility. With architectural and training improvements, recent VLMs can demonstrate strong physical reasoning ability even at smaller scales.

Nevertheless, our tuned PhyDetEx substantially outperforms all competitors across Acc. Impl., F1 Score, and Reasoning Score on both datasets, achieving state-of-the-art results. PhyDetEx does not reach the highest Acc. Plaus., which is largely due to current VLMs' strong bias toward predicting plausibility. Therefore, the F1 Score is the most appropriate and balanced measure for this task. Moreover, the superior Reasoning Score confirms that PhyDetEx can not only detect physical implausibility but also provide accurate, interpretable explanations leveraging the VLM's inherent reasoning capabilities. We also provide performance analysis of the baseline in \textbf{Supplemental Material E}.

\subsection{Case Study} 
\begin{figure*}[!t]
\captionsetup{skip=4pt}
  \centering
   \includegraphics[width=1\linewidth]{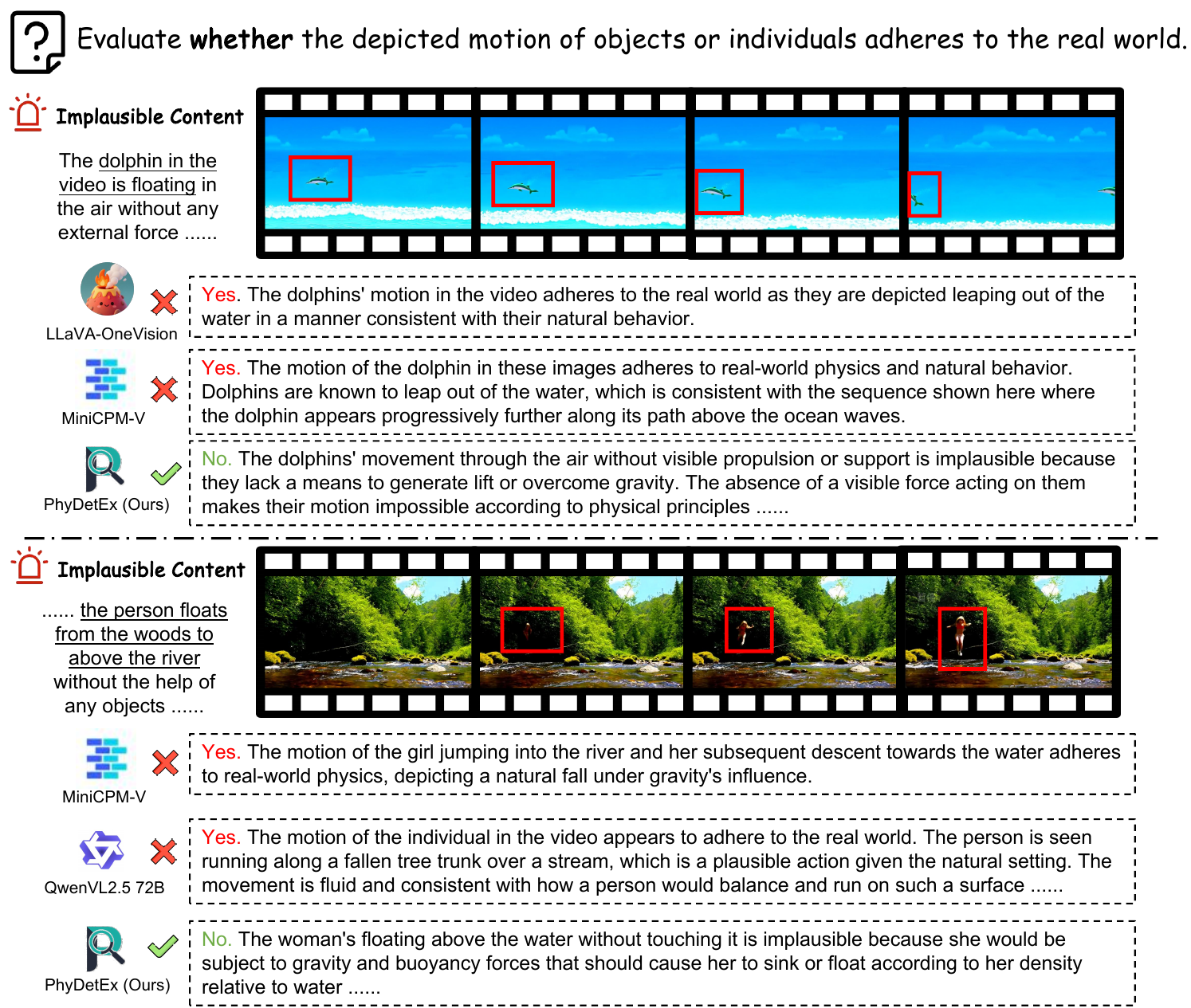}
   \caption{Qualitative comparison between our PhyDetEx and recent VLMs on detecting physical implausibility. We illustrate two representative cases from our PID test split. In the first case (top), the dolphin remains floating above the sea surface without descending, violating gravity. LLaVA-OneVision and MiniCPM-V misinterpret the scene as a normal leap. By contrast, our PhyDetEx correctly identifies the physical implausibility and attributes it to the lack of gravitational influence. In the second case (bottom), a woman jumps from the woods toward the river but remains suspended midair. Other VLMs again regard this as plausible or irrelevant to physics, whereas our PhyDetEx identifies the implausible subject and provides the correct reasoning that her motion defies gravity and buoyancy.}
   \label{fig:casestudy}
\end{figure*}

In Fig.~\ref{fig:casestudy}, we present two representative examples from our PID test split, comparing the outputs of PhyDetEx with several recent VLMs.
The first case depicts a dolphin floating above the sea surface. LLaVA-OneVision and MiniCPM-V misinterpret this as a dolphin jumping out of the water, thus considering it physically plausible. They fail to recognize that the dolphin's height remains nearly constant over time, showing no effect of gravity. In contrast, PhyDetEx correctly identifies the scene as physically implausible and explicitly reasons that the object does not fall under gravity.

The second case is more complex: a woman jumps from a jungle cliff toward a river's surface but remains suspended above it without descending. Similar to the first example, MiniCPM-V again overlooks the implausibility and interprets the scene as a normal jumping motion, while QwenVL2.5-72B focuses on general scene description rather than the physical implausibility. In comparison, PhyDetEx successfully identifies the implausible subject and provides a concise, accurate explanation for its abnormality.
We also provide additional qualitative examples and detailed discuss the failure cases with the corresponding reasoning in \textbf{Supplemental Material F}.

\subsection{Benchmarking Modern T2V Models}

\begin{table}[t]
\centering
\captionsetup{skip=4pt}
\resizebox{0.9\columnwidth}{!}{

\begin{tabular}{lrc}
\toprule
\multicolumn{1}{c}{\multirow{2}[4]{*}{T2V Model}} & \multicolumn{2}{c}{Phy plausible Evaluation} \\
\cmidrule{2-3}          & \multicolumn{1}{c}{Rate} & \multicolumn{1}{c}{\textit{Score}} \\
\midrule
\multicolumn{3}{c}{\textit{Open-Source}} \\
\midrule
CogVideoX1.5 &  52.0   & 5.35 \\ 
HunyuanVideo &  60.0   & 8.73 \\ 
WanX2.1 1.3B &  66.0  & 16.54 \\ 
WanX2.1 14B & 62.0  & 10.84 \\ 
WanX2.2 14B &  66.0  & 12.96 \\ 
\midrule
\multicolumn{3}{c}{Clsoed-Source} \\
\midrule
Veo3.1 &  72.0   & 12.98 \\ 
Sora2.0 &  86.0   & 32.62 \\ 
\bottomrule
\end{tabular}%

}
\caption{Physical plausibility evaluation results of videos generated by various T2V models, benchmarked with our PhyDetEx. The closed-source model Sora2.0 shows the best performance, significantly higher than the second best player, Veo3.1.}
\label{tab:t2v_res}
\end{table}

Having valiadted that our PhyDetEx can effectively detect the physically implausibility in generated videos, we leverage it to assess the ability of recent state-of-the-art T2V models in generating physically plausible videos. Specifically, we first extract 50 prompts involving clear physical interactions from VIDGEN-1M~\cite{tan2024vdgen-1m}. These prompts are then used to generate videos with a variety of T2V models, including open-source models—\textit{CogVideoX1.5 5B}~\cite{yang2024cogvideox}, \textit{HunyuanVideo}~\cite{kong2024hunyuanvideo}, \textit{WanX2.1 1.3B}, \textit{WanX2.1 14B}, and \textit{WanX2.2 14B}~\cite{wan2025}—as well as the  closed-source systems—\textit{Veo3.1}~\cite{Veo3Google} and \textit{Sora2.0}~\cite{sora_2}. Finally, we apply PhyDetEx to detect whether the generated videos exhibit physical implausibility. 

PhyDetEx produces not only binary judgments but also token-level logits. Based on these outputs, we compute two complementary indicators: (1) the \textbf{Plausibility Rate}, defined as the proportion of videos judged to be physically plausible; and (2) a fine-grained \textbf{Plausibility Score} derived from the logits corresponding to the plausibility prediction, providing a continuous measure of physical realism. Detailed information on the benchmarking procedure for T2V models is provided in the \textbf{Supplemental Material G}. In particular, for videos judged as ``Physically Plausible'' by PhyDetEx, the corresponding logits are added to the score, whereas for videos judged as ``Physical Implausible'', the logits are subtracted, resulting in an overall cumulative score.

The results are shown in Tab. \ref{tab:t2v_res}. We see that the closed-source T2V system Sora2.0 achieves significantly higher \textit{Rate} and \textit{Score} than other models, demonstrating its superior understanding of physical laws. Veo3.1 is the second best but lags much behind Sora2.0. 
Among open-source models, \textit{WanX2.1 1.3B} attains the same \textit{Rate} as \textit{WanX2.2 14B} and achieves the highest \textit{Score}. Our analysis indicates that although \textit{WanX2.1 1.3B} produces lower visual quality compared to larger models, this does not reflect a deficiency in its physical reasoning capability. Rather, its tendency to generate simpler actions naturally reduces the likelihood of producing physically implausible content.

Besides benchmarking existing T2V models, we also leverage PhyDetEx to generate positive and negative samples for these models, enabling physical awareness Direct Preference Optimization (DPO~\cite{rafailov2023direct}) to improve them. Detailed implementation and results are provided in \textbf{Supplemental Material H}.

\section{Conclusion}
\label{sec:conclusion}
In this work, we first investigated why current VLMs struggled to identify physical implausibility in generated videos. We then proposed to construct PID, a dataset specifically designed for the physical implausibility detection task, consisting of a train and a test splits. Based on the train split, we developed PhyDetEx, a VLM-based detector capable of not only identifying but also explaining physical implausibility. PhyDetEx achieved SOTA performance on both our test split and the public Impossible Video benchmark. Furthermore, leveraging PhyDetEx, we evaluated a range of modern T2V models in terms of their ability to generate physically plausible videos. Our findings revealed that the powerful closed-source models have made significant progress in producing physically consistent videos, but the open-source models still lagged substantially behind.

\vspace{+2mm}
\noindent\textbf{Limitations}. Although our trained  PhyDetEx model performs well in detecting physical implausibility in generated videos, the diversity of physical events is vast. Therefore, PhyDetEx still exhibits hallucinations in non-visual or complex physical events. We provide several failure cases and discussion of PhyDetEx in \textbf{Supplemental Material F}. We believe that adding more modality information from the video, such as audio and additional information about objects in the video, rather than relying solely on purely visual modalities, can better help VLM understand and detect physical implausibility.

\bibliographystyle{plainnat}
\bibliography{main}

\newpage
\clearpage
\appendix
\appendix

\clearpage
\onecolumn        
\setcounter{page}{1}

\begin{center}
    {\Large 
    \textbf{PhyDetEx: Detecting and Explaining the Physical Plausibility of T2V Models} \\[0.5em]
    -- Supplementary Material --
    }
\end{center}
\vspace{1em}

\setcounter{section}{0}
\setcounter{table}{0}
\setcounter{figure}{0}
\setcounter{footnote}{0}

\renewcommand{\thefootnote}{\arabic{footnote}}
\renewcommand\thesection{\Alph{section}}
\renewcommand\thefigure{S\arabic{figure}}
\renewcommand\thetable{S\arabic{table}}

\noindent\textbf{We highly recommend watching the supplementary video, which comprehensively demonstrates our motivation and results, building a good starting point to understand our work.}

\section{Details of Preliminary Investigation}
\label{sub:detials_of_prelimeinary_exp}


\begin{table*}[h]

\begin{center}
\begin{tcolorbox}[
  colframe=blue!20!black,    
  colback=gray!5,           
  coltitle=black,              
  colbacktitle=blue!20,        
  title={Prompts with Different Conditions}
]
\renewcommand{\arraystretch}{1.3}
\begin{tabularx}{\textwidth}{lX}
\toprule
\rowcolor{green!3}
\textbf{Prompt (C1)} & Given the video content, evaluate whether the depicted motion of objects or individuals adheres to the real world. Start your answer with Yes. or No. and give the explanation. \\
\midrule
\rowcolor{green!12}
\textbf{Prompt (C2)} & Given the video content, \textbf{which may be generated by artificial intelligence}, which is generated by artificial intelligence, evaluate whether the depicted motion of objects or individuals adheres to the real world. Start your answer with Yes. or No. and give the explanation. \\
\midrule
\rowcolor{green!30}
\textbf{Prompt (C3)} & Given the video content, \textbf{which is generated by artificial intelligence}, evaluate whether the depicted motion of objects or individuals adheres to the real world. Start your answer with Yes. or No. and give the explanation. \\
\bottomrule
\end{tabularx}
\end{tcolorbox}
\end{center}
\caption{Prompts used under the preliminary investigation conditions. C1 provides no information about the video source, C2 weakly suggests that the video may be AI-generated, and C3 explicitly states that the video is AI-generated.}
\label{tab:prompt_in_pre_invest}
\end{table*}

\noindent\textbf{Prompts under three conditions.}
Tab.~\ref{tab:prompt_in_pre_invest} lists all prompts used in our preliminary investigation. The prompts under different conditions provide the VLM with varying degrees of information regarding whether the given video originates from an AIGC model.


\section{Ablation Study}
\label{sub:detials_of_training_phydetex}
In this section, we examine whether the design of our PID training split and the training strategy effectively enhance the VLM's ability to understand physical plausibility.

Tab.~\ref{tab:ablation_study} reports the results.  Specifically, we first randomly sample 2,000 paired videos from the PID training split and train a new baseline model (denoted as \emph{Baseline*}) using the same training configuration described in Sec.~5 of the main paper. Based on this randomly selected subset, we then construct two variants:
(1) \textbf{w/o Balanced Data}, where we train the model using only the negative samples; and
(2) \textbf{w/o Paired Data}, where part of the samples are replaced with non-paired videos.
Note that the full PID training split contains 2,588 paired videos; we intentionally select only 2,000 videos here to isolate and validate the role of paired data in the training process.

The results show that removing either the balanced sampling or the paired design leads to a clear performance degradation in F1 Score compared with \emph{Baseline*}. This demonstrates that learning from paired and balanced data is crucial for aligning the model with the distribution of physical plausibility in real world videos.

Furthermore, Tab.~\ref{tab:training_objectives} shows that PhyDetEx with LoRA outperforms full SFT and DPO under the same epochs, suggesting SFT disrupts pretrained knowledge while DPO needs more data.

\begin{table}[t]
\centering
\captionsetup{skip=4pt}
    \begin{tabular}{lcc}
\toprule
Method & Imposs. & PID \\
\midrule
PhyDetEx (LoRA) & \textbf{87.8} & \textbf{83.5} \\
Full SFT & 85.5 & 80.0 \\
DPO & 70.9 & 77.3 \\
\bottomrule
\end{tabular}
\caption{Performance with Different Tuning Strategies}
\label{tab:training_objectives}
\end{table}

\begin{figure*}[t]
\captionsetup{skip=4pt}
  \centering
   \includegraphics[width=0.7\linewidth]{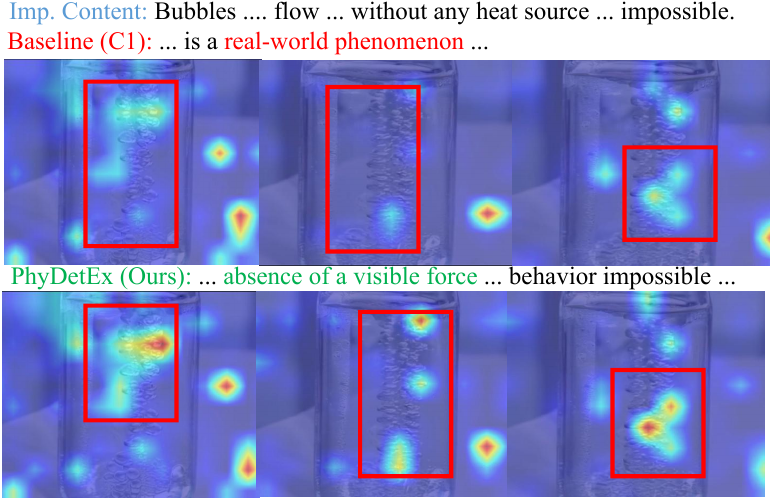}
   \caption{Baseline vs. PhyDetEx attention. PhyDetEx highlights abnormal regions (water, bubbles); baseline is more dispersed.}
   \label{fig:attention}
\end{figure*}

Beyond ablations on data design and training strategy, we further use attention maps to examine whether fine-tuning on the PID training split mitigates shortcut behavior. 
As shown in Fig.~\ref{fig:attention}, the attention maps provide a direct diagnostic signal: after fine-tuning, PhyDetEx focuses more on the abnormal regions, such as water and bubbles, than the baseline on the same example.

\begin{table}[b]
\centering
\captionsetup{skip=4pt}
\resizebox{0.5\columnwidth}{!}{
    \begin{tabular}{l ccc}
\toprule
\multirow{2}{*}{\textbf{Method}} & \multicolumn{1}{c}{\textbf{Acc.}} & \multicolumn{1}{c}{\textbf{Acc.}} & \multicolumn{1}{c}{\textbf{F1}} \\ 
& \textbf{Impl.} & \textbf{Plaus.} & \textbf{Score} \\
\midrule
Base Model & 64.9 & 69.8 & 70.3 \\
\midrule
Baseline* &98.3&  70.8 & 82.2\\
w/o balanced Data & 100.0& 0.0&0.0 \\
w/o paired Data &2.4 & 99.8&71.3 \\
\midrule
PhyDetEx & 86.4&87.1& 87.8 \\
\bottomrule
\end{tabular}
}
\caption{Ablation study on data design and training strategy. We compare the base model, the reconstructed Baseline*, and variants without balanced data or paired data. Both balanced sampling and paired supervision are essential for achieving high physical plausibility detection performance, as reflected in the F1 Score. }
\label{tab:ablation_study}
\end{table}

\section{Details of PID}
\label{sub:detials_of_phyd_2}
In this section, we provide the detailed statistics of our proposed PID dataset.

\vspace{+1mm}
\noindent\textbf{PID training split}. The PID training split consists of two components: (1) positive videos extracted from VIDGEN-1M, which are real-world videos without any physical implausibility, and (2) negative videos generated from captions whose physical content has been intentionally modified to introduce physical implausibility. Specifically, we use HunyuanVideo to synthesize the negative videos from these edited captions. Leveraging our findings from the preliminary investigation, we additionally provide the VLM with the information that each generated video is produced by an AIGC model, and employ Qwen2.5-VL 72B to filter out any generated videos that fail to exhibit the intended physical implausibility.

\vspace{+1mm}
\noindent\textbf{PID test split}. As described in Sec.~4 of the main paper, the construction of the PID test split is designed to more accurately evaluate a VLM's ability to understand physical plausibility based on two key considerations.

First, all physically implausible samples (\ie, negative samples) should originate from \emph{normal prompts}. This design choice aligns with real-world scenarios, where detecting physical implausibility in generated videos typically does not come from the prompts with physical implausibility.
Second, the physically plausible set should include both generated videos and real-world videos. If all physically plausible videos are sourced exclusively from real-world videos, as done in datasets such as Impossible Video, the test split would introduce a shortcut based on ``generated vs.~non-generated'', preventing a fair assessment of whether the VLM truly understands physical plausibility rather than merely detecting generation artefacts.

\begin{figure*}[t]
\captionsetup{skip=4pt}
  \centering
   \includegraphics[width=1\linewidth]{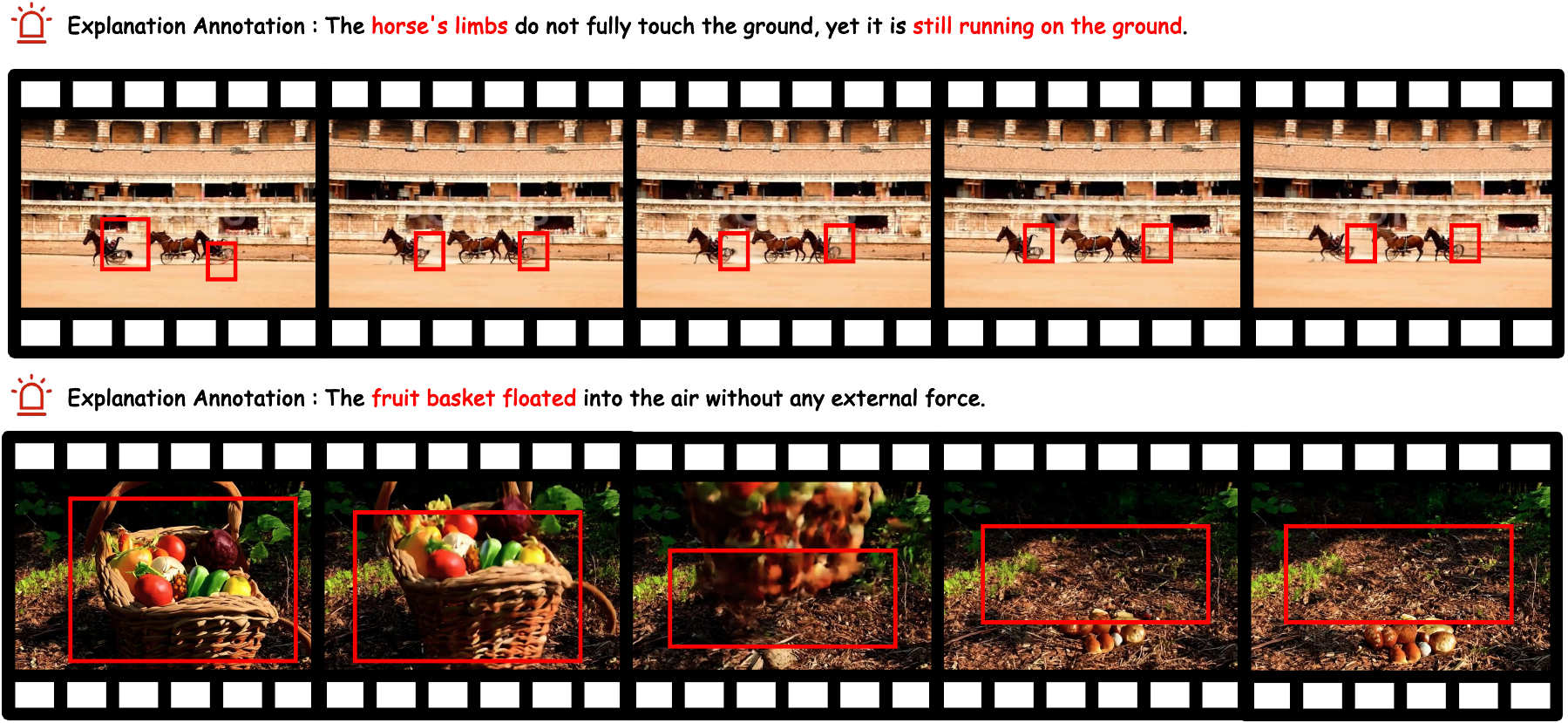}
   \caption{Samples in our PID test split.}
   \label{fig:pid_samples}
\end{figure*}

Following these principles, we manually select physically implausible videos from WanX2.1-1.3B and CogVideoX (2B, 5B) and annotate them accordingly. Similarly, we collect generated and real-world videos without physical implausibility to form the positive samples. Together, these constitute the PID test split. Detailed statistics are provided in Tab.~\ref{tab:pid_test_static}.
We also provide some samples from our PID test split in Fig.~\ref{fig:pid_samples}.

\begin{table}[b]
\captionsetup{skip=4pt}
\centering
\resizebox{0.5\columnwidth}{!}{
    \begin{tabular}{llc}
\toprule
\textbf{Category} & \textbf{Source / Model} & \textbf{Count} \\
\midrule
\multirow{3}{*}{Positive Samples} & Real World & 148 \\
 & Generated (General) & 102 \\
 \cmidrule(l){2-3}
 & \textit{total} & \textbf{250} \\
\midrule
\multirow{4}{*}{Negative Samples} & Wanx2.1 1.3B & 195 \\
 & CogVideoX-5b & 28 \\
 & CogVideoX-2b & 27 \\
 \cmidrule(l){2-3}
 & \textit{total} & \textbf{250} \\
 & \textit{Avg. Explanation} & \textbf{27.58} words\\
\midrule
\textbf{Total} & & \textbf{500} \\
\bottomrule
\end{tabular}
}
\caption{Statistics of the collected PID test split. The total number of videos is 500, evenly split into positive and negative samples. The average annotation length is 27.58 words.}
\label{tab:pid_test_static}
\end{table}

\section{Details of Evaluation}
\label{sub:detials_of_evaluation}

\begin{table}[b]
\begin{center}
\begin{tcolorbox}[
  colframe=blue!20!black,    
  colback=gray!10,           
  coltitle=black,              
  colbacktitle=blue!20,        
  title={Prompt}
]
You will be given a ground truth and a model output couple.\\
Your task is to provide a 'total rating' scoring how well the model output matches the semantic meaning of the ground truth.\\
Give your answer as an integer on a scale of 0 to 5, where 0 means that the model output is completely unrelated to the ground truth, and 5 means that the model output perfectly matches the semantic meaning of the ground truth.\\

Provide your feedback as follows:\\

Feedback:\\
Total rating: (your rating, as an integer between 0 and 5)\\

Now here are the ground truth and model output.\\

Ground Truth: \texttt{\{ground\_truth\}}\\
Model Output: \texttt{\{model\_output\}}\\

Feedback:\\
Total rating: \\

\end{tcolorbox}
\end{center}
\caption{Prompt used in the evaluation of \textit{Reasoning Score}.}
\label{tab:example_prompt_judge}
\end{table}

In this section, we describe the evaluation protocols for all metrics. For Acc. Impl., Acc. Plaus., and F1 Score, we extract the first token of each model's response—either ``Yes'' or ``No''—using regular expression matching, and use this predicted binary judgment to compute the corresponding classification metrics.

For the Reasoning Score, we employ an LLM (LLaMA-3.1 8B Instruct) as an automatic evaluator to assess the semantic alignment between the model's explanation and the ground-truth rationale. To ensure fairness, we remove the initial judgement token (``yes'' or ``no'') from the response before evaluation, so that the score reflects only the quality of the explanation rather than the predicted label, and the scores range from 0 to 5. The prompts used for this evaluation is provided in Tab.~\ref{tab:example_prompt_judge}. 

\section{More Analysis of Baseline}
To better understand why current powerful VLMs fail to detect physical implausibilities in videos, we further analyse their reasoning quality when their judgments are correct. Specifically, we compute the average reasoning score conditioned on correct predictions, termed the \textbf{C}orrect-judgment \textbf{R}easoning \textbf{S}core (\textbf{CRS}). This metric evaluates whether a model truly understands the physical inconsistency in the video when it produces a correct binary decision—that is, whether its generated reasoning content is itself accurate and physically grounded.

As shown in Tab.~\ref{tab:crs}, Qwen3-VL-30B-A3B, a strong baseline, achieves a higher CRS, indicating a more reliable and physically consistent understanding of the underlying content. This improvement is also reflected in its superior accuracy and related metrics. Conversely, a lower CRS suggests that even when a model outputs the correct judgment, its reasoning may be flawed or superficial, which ultimately correlates with lower overall performance.

\begin{table}[t]
\centering
\captionsetup{skip=4pt}
\resizebox{0.35\columnwidth}{!}{
    \begin{tabular}{cc}
\toprule
Model & CRS \\
\midrule
QwenVL2.5 32B & 3.17 \\
QwenVL2.5 72B & 4.61 \\
Qwen3-VL-30B-A3B & 4.71 \\
\bottomrule
\end{tabular}
}
\caption{The average \textbf{C}orrect-judgment \textbf{R}easoning \textbf{S}core (\textbf{CRS}) for the baseline.}
\label{tab:crs}
\end{table}

\section{More Cases and Failure Cases}
\label{sub:more_cases}
\begin{figure*}[t]
\captionsetup{skip=4pt}
  \centering
   \includegraphics[width=0.85\linewidth]{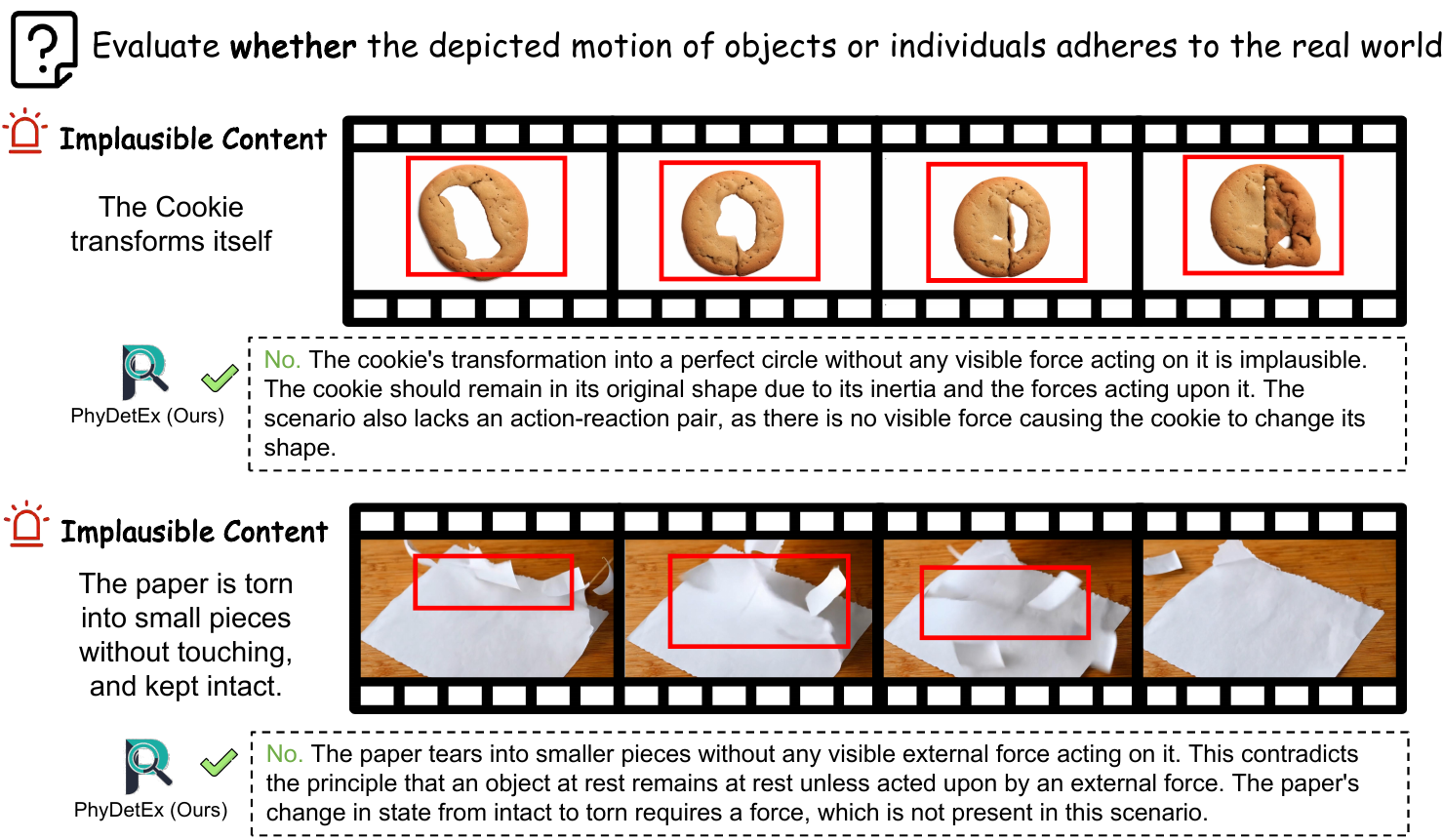}
   \caption{Qualitative examples of PhyDetEx on physically implausible video. Top: a cookie spontaneously transforms its shape without any external force. Bottom: a sheet of paper tears itself into multiple pieces without contact and remains intact afterwards. Both cases violate fundamental physical principles, and PhyDetEx correctly flags them as implausible and gives the corresponding explanation.}
   \label{fig:more_cases}
\end{figure*}

In this section, we present additional qualitative examples of PhyDetEx, along with failure cases that highlight its limitations. Fig.~\ref{fig:more_cases} shows two representative cases correctly identified by our model. In the first video, a cookie spontaneously transforms its shape without any external force, which violates basic physical principles such as inertia and action–reaction. In the second example, a sheet of paper tears itself into smaller pieces without being touched and remains intact after flying apart—another scenario that is clearly physically implausible. PhyDetEx successfully identifies both cases as containing physical implausibility and gives the corresponding explanation.

Although PhyDetEx is effective in detecting physical implausibility in generated videos, it still exhibits several limitations. This is primarily due to the diverse manifestations of physical phenomena, many of which are inherently difficult to assess based solely on visual cues. We categorize its errors into three types: (1) correctly identifying an abnormality but providing an incorrect explanation; (2) incorrectly predicting implausibility for a physically plausible video; and (3) failing to detect an existing physical anomaly.

\begin{figure*}[t]

  \centering
   \includegraphics[width=0.85\linewidth]{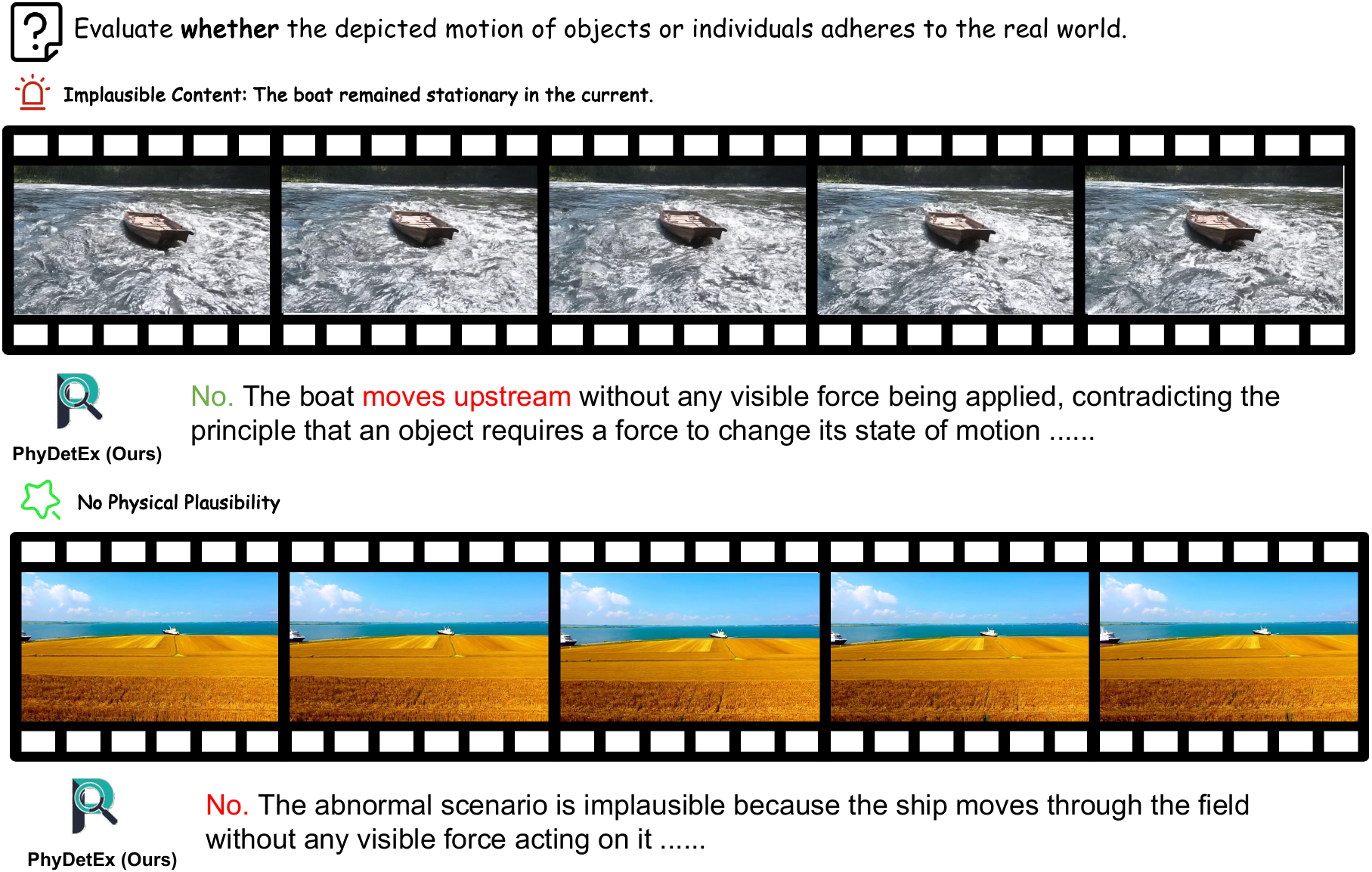}
   \caption{Examples of two failure types. (1) Correctly identifying physical implausibility but providing an incorrect explanation; and (2) incorrectly predicting implausibility for a physically plausible video. Top: The model correctly flags the video as implausible but generates an erroneous explanation, claiming that the boat moves upstream instead of recognizing that it remains stationary despite the strong water current. Bottom: A physically plausible video that PhyDetEx incorrectly classifies as implausible, accompanied by a mistaken explanation that the boat is moving across the field.}
   \label{fig:failure_cases}
\end{figure*}

In Fig.~\ref{fig:failure_cases}, we illustrate the first two categories. In the first example, PhyDetEx correctly determines that the video contains physically implausible motion, but offers an incorrect explanation. The actual implausibility stems from the boat remaining stationary despite a strong water current; however, PhyDetEx erroneously claims that the boat is moving upstream. In the second example, the video is physically plausible, yet PhyDetEx incorrectly predicts implausibility and generates an erroneous explanation, mistakenly assuming that the boat is moving across the field.

\begin{figure*}[t]
  \centering
   \includegraphics[width=0.85\linewidth]{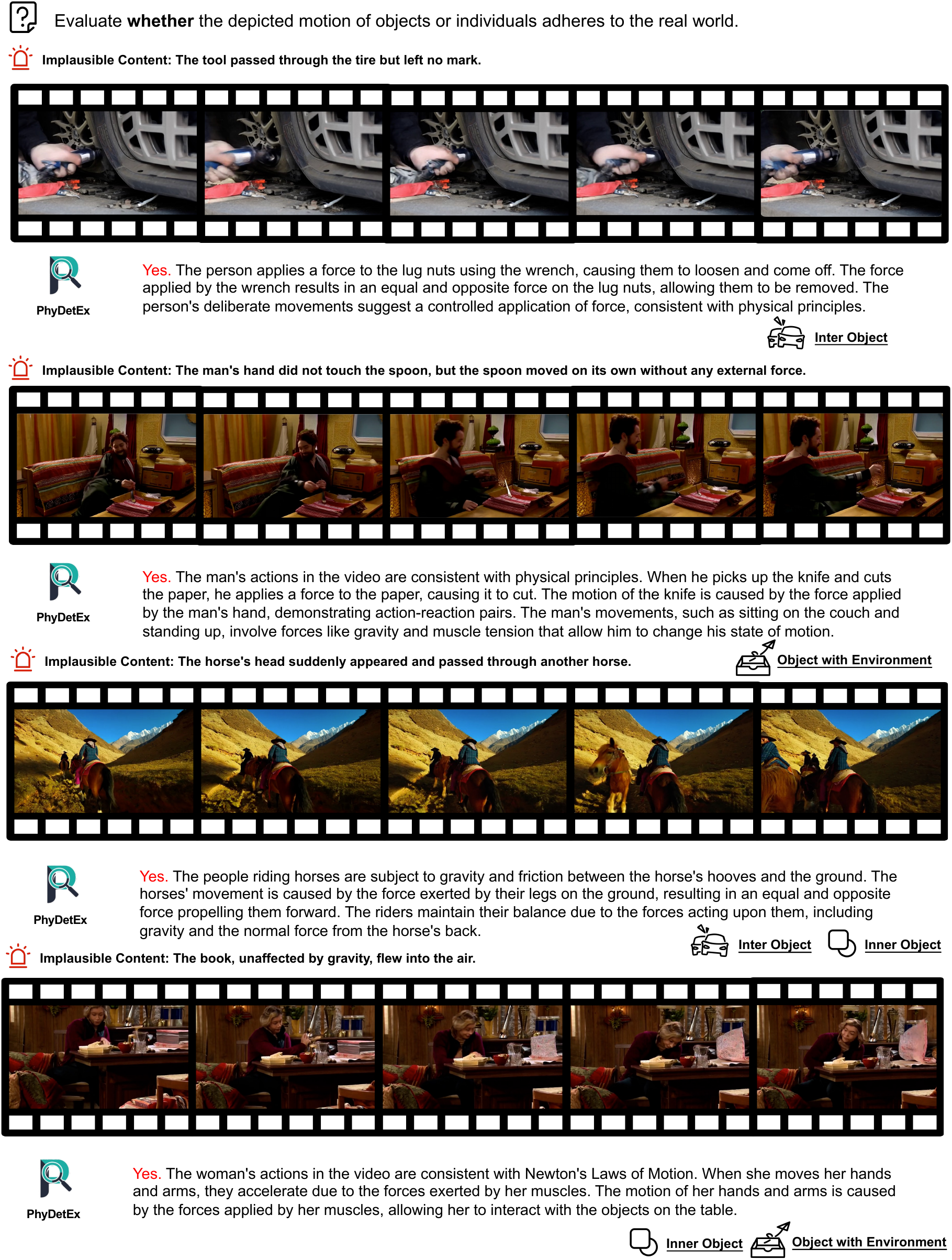}
   \caption{Examples of the third failure type, i.e., failure to detect existing physical anomalies. The errors are categorized according to their visual manifestation into three groups: \textit{Inner-Object} (e.g., deformation), \textit{Inter-Object} (e.g., collision), and \textit{Object–Environment} interactions (e.g., gravity-related inconsistencies).}
   \label{fig:failure_cases_by_type}
\end{figure*}

For the third type, missed detection of existing anomalies, we further categorize the errors based on their visual manifestation into three groups: \textit{Inner-Object} (e.g., object deformation), \textit{Inter-Object} (e.g., collisions between objects), and \textit{Object with Environment} (e.g., gravity-related inconsistencies). As shown in Fig.~\ref{fig:failure_cases_by_type}, many physical laws in this category are not directly observable from appearance alone, particularly for Inner-Object and Object–Environment cases. This poses a fundamental challenge for methods like PhyDetEx that rely exclusively on video inputs. It is also worth noting that, in some instances, multiple types of physical anomalies may co-occur within the same video.

\section{Benchmarking the T2V Models}
\label{sub:rank_the_t2v}

\subsection{Benchmarking Details}
Using the tuned PhyDetEx, we evaluate a variety of current open-source and closed-source T2V models. For each video generated from prompts extracted from VIDGEN-1M, PhyDetEx determines whether the video contains physically implausible content. We extract the model's binary judgment (\eg, Yes or No) using a regular expression applied to the first token of its output, which forms the basis of the \textit{Rate} metric.
To obtain a more fine-grained assessment, we further leverage the VLM-based design of PhyDetEx and directly read the logits associated with its judgment, forming the \textit{Score} metric.
It is important to note that the two metrics are not strictly equivalent. For example, although WanX2.1-14B yields a much lower \textit{Rate} compared with Veo-3.1, the difference between the two models becomes marginal under the \textit{Score} metric. This phenomenon arises because some physical events in videos are inherently ambiguous, causing PhyDetEx to produce judgments with varying degrees of confidence.

\subsection{Assessment of T2V Models}

\textbf{Open-Source T2V Models}. All experiments involving open-source T2V models are conducted on servers equipped with NVIDIA A800 GPUs. The models evaluated in this work are obtained from the following sources:
\begin{itemize}
\item \textbf{CogVideoX-1.5}: downloaded from the official GitHub repository.\footnote{\url{https://github.com/THUDM/CogVideo}}

\item \textbf{Wan2.1-14B and Wan2.1-1.3B}: accessed via the official GitHub repositories,\footnote{\url{https://github.com/Wan-Video/Wan2.1}}
 with checkpoints and inference scripts from Hugging Face.\footnote{\url{https://huggingface.co/Wan-AI}}

\item \textbf{Wan2.2-14B}: accessed via the official GitHub repository,\footnote{\url{https://github.com/Wan-Video/Wan2.2}}
 with checkpoints and inference scripts from Hugging Face.\footnote{\url{https://huggingface.co/Wan-AI/Wan2.2-T2V-A14B}}

\item \textbf{Hunyuan Video}: downloaded from the official GitHub repository.\footnote{\url{https://github.com/Tencent-Hunyuan/HunyuanVideo}}

\end{itemize}

\vspace{+1mm}
\noindent\textbf{Closed-Source T2V Models}. Closed-source models are accessed exclusively through their official APIs: Veo-3.1 via the Google Gemini API\footnote{\url{https://developers.googleblog.com/en/veo-3-now-available-gemini-api/}}
 and Sora-2 via the OpenAI API.\footnote{\url{https://api.openai.com/v1/videos}}

Due to the high cost of video generation for both Veo-3 and Sora-2, our experiments rely on the Veo-3.1-Fast and Sora-2-Standard versions.

\section{Implementation of Physical-Aware DPO}
\label{sub:physical_aware_dpo}

Beyond using PhyDetEx as an evaluator for existing T2V models, we also employ it as a data generator for DPO training. Specifically, PhyDetEx provides comparative judgments across multiple videos generated from the same prompt, enabling us to construct positive–negative preference pairs for Direct Preference Optimization (DPO). Fine-tuning T2V models on such physically grounded preference data improves their ability to produce physically plausible videos. The overall procedure is described below.

\vspace{+1mm}
\noindent\textbf{Collecting DPO Data.}
We begin by sampling a large set of prompts from the VidProM dataset, which is derived from real user communities. Low-quality prompts or those lacking meaningful physical interactions are filtered using an LLM. For each remaining prompt, the T2V model generates 12 videos, which are then evaluated by PhyDetEx.
Because the 12 videos may contain multiple plausible samples or multiple implausible samples, we rely on the evaluation logits to determine the most confident positive and negative instances. If both positive and negative samples are present, we select the highest-logit plausible video as the positive sample, and the highest-logit implausible video as the negative sample.
If one category is absent (\eg, no plausible video), we use the lowest-logit video of the opposite category as a surrogate to ensure that each prompt yields a paired sample. 

\vspace{+1mm}
\noindent\textbf{Fine-Tuning the T2V Model.}
With the DPO dataset, we conduct experiments using WanX2.1-1.3B. We implement the standard DPO loss and additionally incorporate an MSE reconstruction loss to preserve the model's basic generation quality and prevent visual collapse. In our configuration, the DPO loss weight is set to 0.5 and the MSE loss weight to 1.0.
We train the model for one epoch on 2,500 preference pairs generated by WanX2.1-1.3B itself, with a learning rate of 1e-5 and a cosine Learning Rate scheduler.

\vspace{+1mm}
\noindent\textbf{Evaluation.}
Following fine-tuning, we evaluate the model on VideoPhy, which is external to the PhyDetEx training domain. As shown in Tab.~\ref{tab:dpo_res}, Physical-Aware DPO yields consistent gains on both Physical Commonsense (PC) and Semantic Alignment (SA). For comparison, we additionally conduct DPO using a vanilla VLM (QwenVL2.5 7B, the backbone of PhyDetEx) to construct preference data. Although this approach improves semantic alignment, the vanilla VLM’s weaker physical reasoning capability results in degraded performance on physical commonsense, highlighting the importance of physics-aware supervision.

\begin{figure*}[t]
  \centering
   \includegraphics[width=0.6\linewidth]{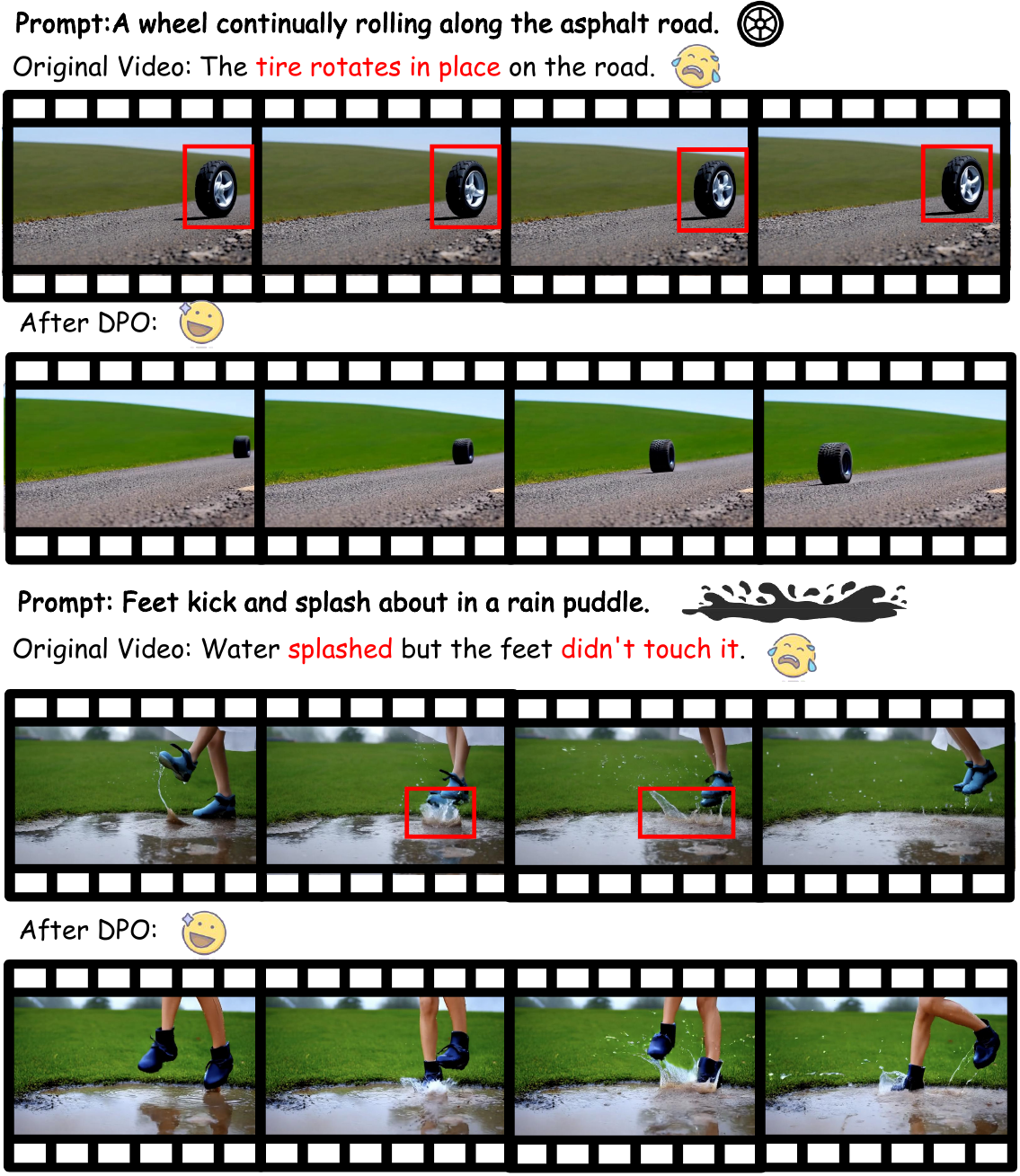}
   \caption{Qualitative examples of results after DPO fine-tuning.}
   \label{fig:dpo_cases}
\end{figure*}

Qualitative examples from VideoPhy are provided in Fig.~\ref{fig:dpo_cases}. In the first case, a tyre previously rotated in place without moving downward along the slope; after DPO training, the tyre rolls naturally along the incline. In the second case, a foot produced a water splash despite not contacting the water surface—an obvious physical violation. After DPO training, the splash appears only upon actual contact, aligning with real-world physical behaviour.
Together, these results demonstrate not only the reliability of PhyDetEx in detecting physical implausibilities but also its wide utility in downstream T2V model enhancement.

\begin{table}[t]
\centering
\captionsetup{skip=4pt}
\resizebox{0.6\columnwidth}{!}{
    \begin{tabular}{lcc}
\toprule
 \textbf{Method}  & PC    & SA \\
\midrule
Baseline & 36.16 & 42.66 \\
DPO(vanille) & 35.92(-0.24) & 42.69(+0.03) \\
DPO(PhyDetEx) & \textbf{37.63(+1.47)} & \textbf{43.81(+1.15)} \\
\bottomrule
\end{tabular}%

}
\caption{Performance of WanX2.1 1.3B before and after Physical-Aware DPO. We further report results obtained using DPO data constructed with a vanilla VLM (QwenVL2.5 7B). DPO fine-tuning guided by PhyDetEx consistently improves performance across both metrics on the VideoPhy benchmark.}
\label{tab:dpo_res}
\end{table}

\end{document}